# Optimizing Transformer based on high-performance optimizer for predicting employment sentiment in American social media content

Feiyang Wang[1*], Qiaozhi Bao[2], Zixuan Wang[3], Yanlin Chen[4],
[1]Information Networking Instutite, Carnegie Mellon University, California, Sunnyvale, 94085, USA.
[2]Department of Statistics, North Carolina State University, North Carolina, Raleigh, 27695, USA.
[3]College of Engineering, Carnegie Mellon University, California, Mountain View, 94035, USA.
[4]Tandon School of Engineering, New York University, New York, Brooklyn, 11201 , USA.
[*]Corresponding author: e-mail: feiyangw@alumni.cmu.edu

*Abstract*—This article improves the Transformer model based on swarm intelligence optimization algorithm, aiming to predict the emotions of employment related text content on American social media. Through text preprocessing, feature extraction, and vectorization, the text data was successfully converted into numerical data and imported into the model for training. The experimental results show that during the training process, the accuracy of the model gradually increased from 49.27% to 82.83%, while the loss value decreased from 0.67 to 0.35, indicating a significant improvement in the performance of the model on the training set. According to the confusion matrix analysis of the training set, the accuracy of the training set is 86.15%. The confusion matrix of the test set also showed good performance, with an accuracy of 82.91%. The accuracy difference between the training set and the test set is only 3.24%, indicating that the model has strong generalization ability. In addition, the evaluation of polygon results shows that the model performs well in classification accuracy, sensitivity, specificity, and area under the curve (AUC), with a Kappa coefficient of 0.66 and an F-measure of 0.80, further verifying the effectiveness of the model in social media sentiment analysis. The improved model proposed in this article not only improves the accuracy of sentiment recognition in employment related texts on social media, but also has important practical significance. This social media based data analysis method can not only capture social dynamics in a timely manner, but also promote decision-makers to pay attention to public concerns and provide data support for improving employment conditions.

## I. INTRODUCTION

The research background of social media emotion prediction stems from the rapid development of information technology and the popularity of social platforms. With the rise of social media platforms such as Facebook, Twitter, and Instagram, the frequency of users sharing personal opinions and emotions on these platforms has significantly increased. These platforms have not only become important tools for people's daily communication, but also important places for social public opinion and emotional expression. Researchers are gradually realizing that extracting emotional information from social media can provide valuable insights for fields such as social sciences, marketing, and public policy. Especially in the economic field, there may be a correlation between emotional fluctuations on social media and economic indicators. Therefore, conducting sentiment analysis on social media content can help understand the public's reactions to specific events or topics, such as the job market, and provide reference for decision-makers.

For the employment situation in the United States, emotional prediction of employment related content on social media has important research significance [3]. On the one hand, social media can real-time reflect the public's views on the economic situation and job market, which often affect consumer confidence and business investment decisions. For example, when a large number of users express their anxiety about employment prospects on social media platforms, this may indicate a decline in consumer spending, thereby affecting overall economic growth. On the other hand, analyzing the emotional tendencies of employment related topics (such as job postings, career development opinions, etc.) can help policy makers identify potential problems in a timely manner and take corresponding measures to alleviate the instability of the job market. In addition, companies can also use these sentiment analyses to optimize recruitment strategies, improve employee satisfaction, and enhance competitiveness.

Machine learning algorithms play a crucial role in predicting emotions on social media. Through natural language processing (NLP) technology, machine learning models can automatically analyze massive text data and extract potential emotional features. These algorithms, such as support vector machines (SVM) [5], random forests [6], deep learning models, etc., are widely used in text classification and sentiment analysis tasks. For example, Convolutional Neural Networks (CNN) [7] and Long Short Term Memory Networks (LSTM) can effectively capture contextual information in text, thereby improving the accuracy of sentiment classification. In addition, machine learning can optimize models by continuously learning new data to better adapt to the dynamically changing social media environment. This technology not only improves data processing efficiency, but also provides a powerful tool for real-time monitoring of social public opinion, enabling researchers and decision-makers to quickly respond to changes in public sentiment and provide scientific basis for various economic activities. This article is based on swarm intelligence optimization algorithm to improve and optimize the transformer model for predicting emotions in employment related text content on American social media.

## II. DATA FROM DATA ANALYSIS

The dataset of this article is selected from an open-source dataset, which contains a total of 4845 pieces of text content about employment on social media. Each piece of content has its corresponding emotional label, which is divided into positive and negative emotions. Four pieces of data were selected for display, and some of the data are shown in Table 1.

TABLE I. SOME OF THE DATA

| Label | Text |
|---|---|
| positive | STORA ENSO , NORSKE SKOG , M-REAL , UPM-KYMMENE Credit Suisse First Boston ( CFSB ) raised the fair value for shares in four of the largest Nordic forestry groups . |
| positive | A purchase agreement for 7,200 tons of gasoline with delivery at the Hamina terminal , Finland , was signed with Neste Oil OYj at the average Platts index for this September plus eight US dollars per month . |
| negative | The international electronic industry company Elcoteq has laid off tens of employees from its Tallinn facility ; contrary to earlier layoffs the company contracted the ranks of its office workers , the daily Postimees reported . |
| positive | With the new production plant the company would increase its capacity to meet the expected increase in demand and would improve the use of raw materials and therefore increase the production profitability . |

### III. TEXT DATA PROCESSING

Text data processing is an important part of natural language processing (NLP). This article performs text preprocessing, feature extraction, and vectorization to convert textual data into numerical data.

#### A. Text preprocessing

The first step is to remove noise by removing unnecessary characters such as punctuation marks, special characters, and extra spaces.

Then convert all text to lowercase to ensure consistency of the same vocabulary. Secondly, segment the continuous text into individual words or phrases. Next, stop using words and remove some common but meaningless words for analysis, such as "yes", "in", "and", which can help reduce noise and improve analysis efficiency. Finally, stem extraction and word form restoration are performed, and the words are restored to their basic form through algorithms, which helps to unify words of different forms.

#### B. Feature extraction

After completing preprocessing, we extract features from the text for subsequent modeling. This article uses the TF-IDF (Term Frequency Across Document Frequency) method for text feature extraction. TF-IDF is an improved version of the feature extraction method that not only considers the frequency (TF) of a word in a certain document, but also its importance (IDF) in all documents. In this way, the impact of common but low information words on the training results of the model can be reduced.

#### C. Vectorization

The final step is to convert the extracted features into numerical data. Firstly, a feature matrix is constructed using the TF-IDF method to create a matrix between the document and the features. Each row represents a document, each column represents a feature, and the values in the matrix are the weights or frequencies of the corresponding features in the document. Then standardization and normalization are carried out. In order to improve model performance, it is necessary to standardize the feature matrix to eliminate the impact of different orders of magnitude.

Through the above steps, textual data can be successfully transformed into numerical data, which can be utilized by various machine learning models.

### IV. METHOD

#### A. Arctic Penguin Optimization

Arctic Penguin Optimization (APO) is an emerging swarm intelligence optimization algorithm designed to simulate the foraging and survival strategies of Arctic puffins in their natural habitats. This algorithm simulates the behavior of puffins searching for food in cold environments and combines a biologically inspired search mechanism to solve complex optimization problems. The basic principle of APO is to achieve global optimal solutions through collaboration and competition among individuals, utilizing information sharing and adaptive adjustment. When foraging, puffins make decisions based on their surrounding environment and the behavior of other individuals, which provides inspiration for algorithms to effectively explore vast solution spaces [8].

The APO algorithm mainly includes initialization, fitness evaluation, position update, and convergence steps. In the initialization phase, a set of puffin positions is randomly generated first, with each position representing a potential solution. Next, evaluate the performance of each individual in the current environment by calculating their positional fitness (i.e., objective function value). Based on fitness values, the algorithm will select individuals with better performance as "leaders" and guide other individuals towards these excellent solutions. Meanwhile, in order to avoid falling into local optima, APO also introduces a certain degree of randomness and mutation mechanism, allowing individuals to explore new regions and thus improve search efficiency. This dynamic adjustment enables the algorithm to balance the relationship between local search and global search, effectively improving its ability to solve complex problems.

In addition, the APO algorithm has good adaptability and flexibility, and can be widely applied to various optimization tasks, including function optimization, combinatorial optimization, constraint optimization, etc. In practical applications, APO has been used to solve problems such as parameter tuning, engineering design, path planning, etc., and has achieved significant results. Compared with traditional optimization methods, APO can not only handle high-dimensional nonlinear problems, but also effectively address multi-objective optimization challenges. This makes it have broad application prospects in industries, transportation, finance, and other fields. In summary, the Arctic puffin optimization algorithm provides a novel and effective methodology for complex optimization problems with its unique bio inspired mechanism and flexible and efficient search strategy.

*B. Transformer*

The core of Transformer is its self attention mechanism, which allows the model to dynamically focus on different parts of the input sequence when processing it. In the self attention mechanism, each input element generates three vectors: Query, Key, and Value. Specifically, for each word in the input sequence, these three vectors are first obtained through linear transformation [9]. Then, attention scores are obtained by calculating the dot product of the query vector and all key vectors, which represent the importance of the current word to other words. Next, these scores are normalized using the Softmax function to form weights, which are then used to weight and sum the value vectors to obtain the representation of the word in context. This process enables the model to effectively capture the relationships between words in a sentence. The model structure diagram of Transformer is shown in Figure 1.

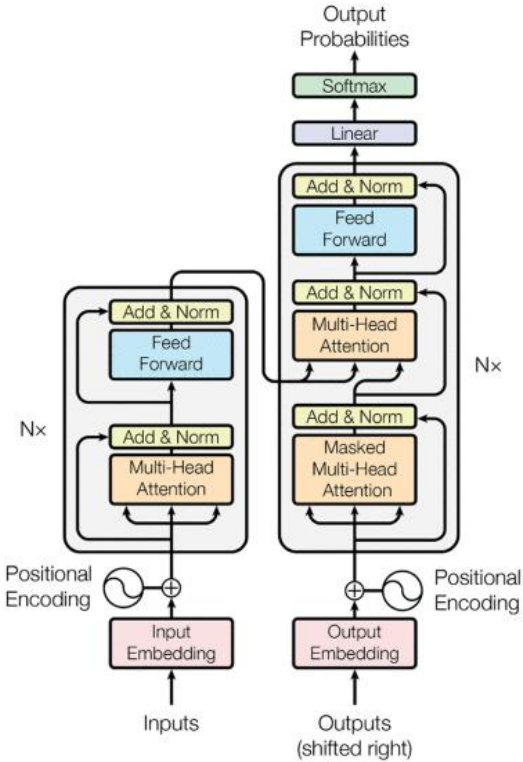

Figure 1. The model structure diagram of Transformer.

The Transformer model consists of an encoder and a decoder, each of which is composed of multiple identical layers stacked together. The encoder is responsible for converting the input sequence into a contextual representation, while the decoder generates the target sequence based on the output of the encoder [10]. Each encoder layer consists of two main components: a multi head self attention mechanism and a feedforward neural network. Multi head self attention allows the model to learn information in parallel in different subspaces, thereby enhancing its expressive power. And feedforward neural networks further process information through nonlinear activation functions. Similarly, the decoder layer also includes a multi head self attention mechanism, but introduces an additional "masking" step to ensure that only the currently generated word can be focused on when generating the next word.

Since Transformer does not use recursive structures, it requires a method to capture the positional relationships of words in the input sequence. For this purpose, location encoding is added to the input embedding to provide location information. These position codes are generated through sine and cosine functions, making them distinguishable between different positions. In addition, due to the ability of Transformer to process all input elements in parallel, compared to traditional RNN structures that gradually process sequences, it significantly improves training speed. This efficiency enables the application of Transformers on large-scale datasets and drives the development of natural language processing.

*C. Optimizing Transformer Based on APO*

The principle of optimizing the Transformer model based on the Arctic puffin optimization algorithm (APO) is mainly reflected in adjusting the hyperparameters and structure of the Transformer through a biologically inspired search mechanism, thereby improving its performance and training efficiency. APO simulates the foraging behavior of Arctic puffins in complex environments, exploring the solution space through collaboration and competition among individuals. When applied to Transformer, this algorithm can optimize multiple key parameters of the model, such as learning rate, hidden layer dimension, attention head count, etc., to achieve better model configuration.

In the specific implementation process, a fitness function is first defined, which is usually based on the performance of the model on the validation set (such as accuracy or loss value). Then, a set of individual puffins is generated using the APO algorithm, with each individual representing a specific combination of hyperparameters. Next, the algorithm will evaluate the fitness of each individual and select outstanding individuals as "leaders" based on their fitness values. Other individuals update by imitating the position of the leader, while introducing a certain degree of randomness to avoid falling into local optima. This dynamic adjustment mechanism enables APO to effectively search for the optimal hyperparameter settings, thereby improving the performance of Transformer models on specific tasks.

V. RESULT

The software used in this experiment is Matlab, the experimental machine is a 3090 graphics card, and the memory is 32GB. In terms of parameter settings, the optimizer selects Adam, the maximum number of training epochs is set to 200, the batch size is set to 256, the initial learning rate is set to 0.0002, the learning rate reduction factor is set to 0.1, and the gradient clipping threshold is set to 10.

Import the model for training, record the values of loss and accuracy during the training process, and output the change curves of loss and accuracy, as shown in Figure 2.

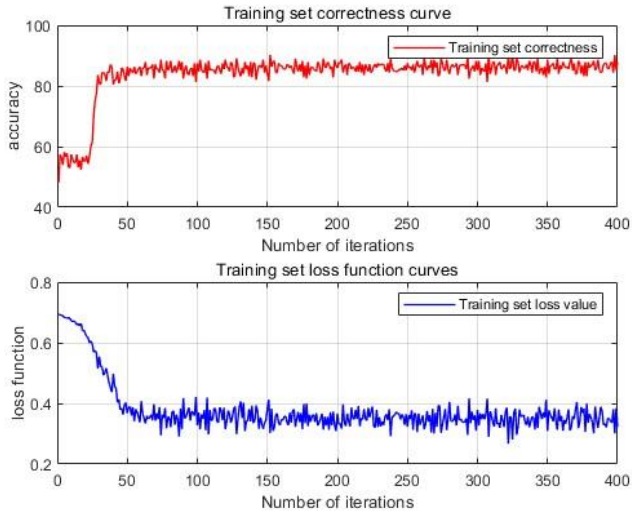

Figure 2. The values of loss and accuracy during the training process.

As shown in the figure, the accuracy of the training set gradually increased from 49.27% at the beginning, to 82.83% at the 28th epoch, and then gradually stabilized and converged. The loss of the training set gradually decreased from 0.67 at the beginning, to 0.35 at the 47th epoch, and then gradually stabilized and converged. The numerical changes in loss and accuracy of the training set indicate that the predictive performance of the model gradually improves on the training set.

Output the confusion matrix of the training set and the test set, observe the difference in the model's predictions between the training set and the test set. The confusion matrix of the training set is shown in Figure 3, and the confusion matrix of the test set is shown in Figure 4.

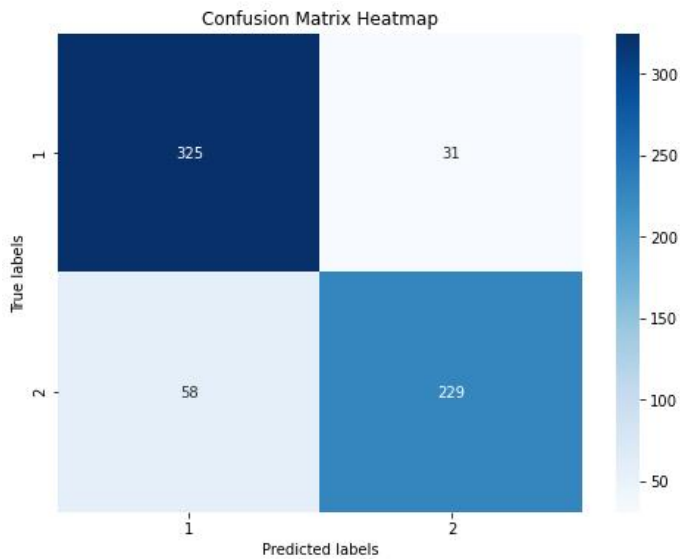

Figure 3. The confusion matrix of the training set.

According to the confusion matrix of the training set, there were 554 correct predictions and 89 incorrect predictions in social media emotion prediction. Among them, 58 positive emotion texts were predicted as negative emotions, and 31 negative emotion texts were predicted as positive emotions. The prediction accuracy of the training set was 86.15%.

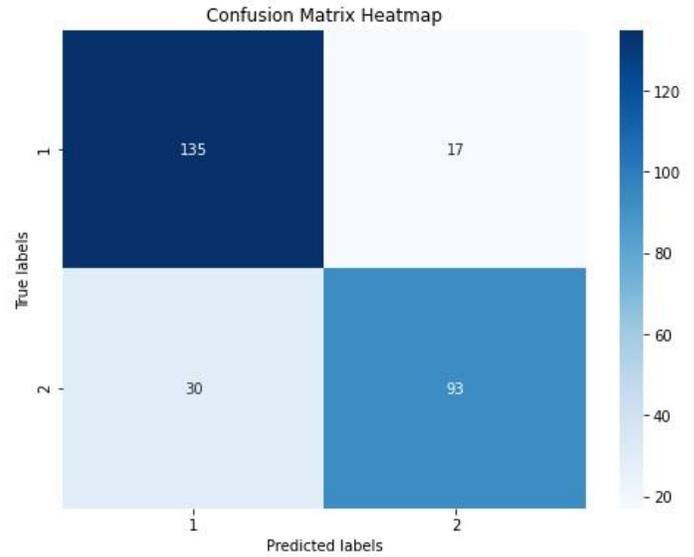

Figure 4. The confusion matrix of the test set.

According to the confusion matrix of the test set, there were 228 correct predictions and 47 incorrect predictions in social media emotion prediction. Among them, 30 texts with positive emotions were predicted as negative emotions, and 17 texts with negative emotions were predicted as positive emotions. The prediction accuracy of the training set was 82.91%.

The accuracy difference between the training set and the test set is 3.24%, which is not significant, indicating that the model proposed in this paper also performs well on the test set and has good generalization ability.

The polygon for predicting and evaluating the output model is shown in Figure 5.

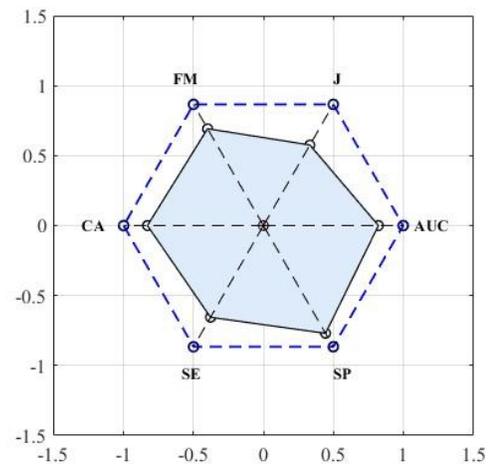

Figure 5. The polygon for predicting and evaluating the output model.

According to the polygon results, the polygon area PAM is 0.63, the classification accuracy is 0.83, the sensitivity is 0.76,

the specificity is 0.89, the area under the curve AUC is 0.82, the Kappa coefficient is 0.66, and the F-measure is 0.80.

## VI. Conclusion

This article focuses on sentiment prediction of employment related text content on American social media, and improves and optimizes the Transformer model based on swarm intelligence optimization algorithm. The research process includes text preprocessing, feature extraction, and vectorization, converting raw text data into numerical data that can be processed by the model. After multiple training sessions, the accuracy of the model on the training set gradually increased from 49.27% to 82.83%, and stabilized after the 28th epoch. Meanwhile, the loss value also decreased from 0.67 to 0.35, indicating that the model gradually converged and improved its prediction performance during the learning process. Through confusion matrix analysis, we found that a total of 554 texts were correctly classified in the predicted results of the training set, while 89 texts were misclassified. Among them, 58 texts with positive emotions were misclassified as negative emotions, and 31 texts with negative emotions were misclassified as positive emotions. The final accuracy of the training set was 86.15%.

On the test set, the model also performed well, with a total of 228 correct predictions and 47 incorrect predictions. Among them, 30 texts with positive emotions were misclassified as negative emotions, and 17 texts with negative emotions were misclassified as positive emotions, resulting in an accuracy rate of 82.91% on the test set. This relatively small difference (3.24%) indicates that the model has good generalization ability and can be effectively applied to unseen data. In addition, through the analysis of polygon results, we obtained a series of important indicators: polygon area PAM of 0.63, classification accuracy of 0.83, sensitivity of 0.76, specificity of 0.89, area under the curve AUC of 0.82, as well as Kappa coefficient of 0.66 and F-measure of 0.80. These results further validate the effectiveness of the proposed method in social media sentiment analysis.

The improved Transformer model proposed in this article not only enhances the accuracy of sentiment analysis of employment related content on American social media, but also helps to better understand the public's attitudes and feelings towards employment issues. In the current economic environment, employment issues have attracted much attention. Through in-depth analysis of relevant discussions on social media, policy makers, enterprises, and job seekers can better grasp the dynamics of social public opinion and provide important references for the job market. This data-driven approach will help promote the formulation and implementation of employment related policies, enhance society's overall understanding and response capabilities to employment issues, and thus promote healthier and more sustainable development.